\documentclass{article}
\usepackage[numbers]{natbib}
\usepackage[preprint]{neurips_2025_custom}
\usepackage[utf8]{inputenc}
\usepackage[T1]{fontenc}
\usepackage{amsmath,amssymb}
\usepackage{graphicx,booktabs,float,xspace,enumitem,microtype,fvextra}
\definecolor{linkColor}{rgb}{0.18,0.39,0.62}
\usepackage[colorlinks=true,linkcolor=linkColor,citecolor=linkColor,
  filecolor=linkColor,urlcolor=linkColor]{hyperref}
\usepackage{fontawesome5}

\newcommand{\ours}{Agensh}
\newcommand{\papertitle}{Agensh: Scaling Organizational Intelligence to 1,024 Agents}
\title{\fontsize{15.5pt}{18pt}\selectfont\papertitle}

\author{
\begin{tabular}{c}
  \\[2pt]
  \textbf{Zhihao Zhan}\thanks{Equal contribution. \textsuperscript{\scriptsize\(\dagger\)} Corresponding authors.}
  ~\;~
  \textbf{Ting Song}\footnotemark[1]
  ~\;~
  \textbf{Li Dong}\textsuperscript{\scriptsize\(\dagger\)}
  \\[2pt]
  \textbf{Shaohan Huang}
  ~\;~
  \textbf{Jianxun Lian}
  ~\;~
  \textbf{Yan Xia}\textsuperscript{\scriptsize\(\dagger\)}
  ~\;~
  \textbf{Furu Wei}\textsuperscript{\scriptsize\(\dagger\)}
  \\[2pt]
  \textnormal{Microsoft Research}
  \\
  \url{https://aka.ms/GeneralAI}
\end{tabular}
}
\AtBeginDocument{\setlength{\headheight}{14.4pt}}
\makeatletter
\renewcommand{\@noticestring}{Technical report.}
\makeatother

\begin{document}
\maketitle
\vspace{-30pt}

\begin{abstract}

\vspace{-4pt}

A multi-agent system can reduce latency on complex tasks by executing work concurrently.
Several pioneering harness frameworks support multi-agent systems.
However, the scalability of current multi-agent harnesses is often constrained by a central
orchestrator's capacity to allocate tasks and coordinate workers. To address this limitation, we
introduce \ours{}, a scalable self-organized multi-agent harness without a central orchestrator:
concurrent workers execute a multi-agent cooperation loop, continuously gathering context,
claiming and self-assigning sub-tasks, taking action and sharing findings, verifying results, and
merging progress in an asynchronous manner.
The loop is supported by the agentic organization infrastructure comprising three components:
a shared workspace holds proposed, ongoing, and completed work; a message interface lets
workers communicate; and shared context retains reusable findings and work intentions. To test
the scalability of \ours{}, we evaluate it on the five hardest ProgramBench tasks with
GPT-5.6-sol (high). Scaling from 1 to 128 agents
raises the mean final test-pass rate from 19.31\% to 28.78\%, an approximately 49\% relative
improvement. Larger organizations reach comparable test-pass rates earlier. On
\texttt{pandoc}, scaling from 1 to 1,024 agents raises the final test-pass rate from 33.89\% to
55.06\%. Worker trajectories further show that different forms of self-organized cooperation gradually
emerges and standardizes as the organization grows. These results reveal the number of agents
as a new scaling dimension for multi-agent
organizations to expand the frontier of general intelligence, offering a practical solution for
complex tasks under hard latency constraints or time budgets.

\vspace{-2pt}

\begin{center}
\begin{tabular}{l}
\faHome \hspace{2pt} \textbf{Project Page:} \url{https://aka.ms/Agensh} \\
\faGithub \hspace{2pt} \textbf{Code:} \url{https://github.com/microsoft/Agensh}
\end{tabular}
\end{center}

\vspace{-16pt}

\end{abstract}

\begin{figure}[H]
  \centering
  \includegraphics[width=\linewidth]{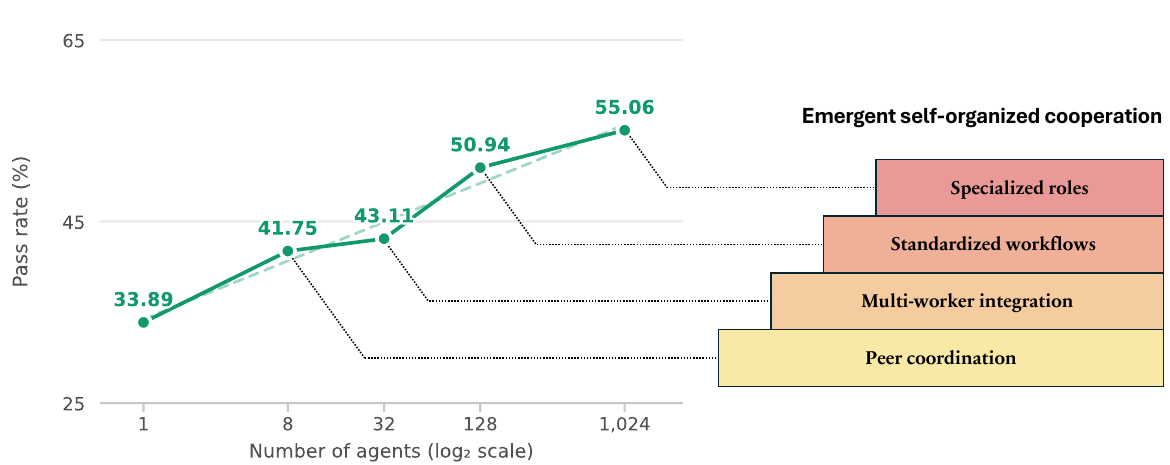}
  \setlength{\abovecaptionskip}{0pt}
  \caption{Scaling from 1 to 1,024 agents on building \texttt{pandoc} \citep{pandoc2026github} from scratch under 6h budget without Internet access.
  Final test-pass rate rises from 33.89\% for 1 agent to 55.06\% for 1,024
  agents. As the organization grows, emergent self-organized cooperation gradually expands from peer coordination
  to multi-worker integration, standardized workflows, and specialized roles.}
  \label{fig:headline-scaling}
\end{figure}

\begin{figure}[H]
  \centering
  \includegraphics[width=0.96\linewidth]{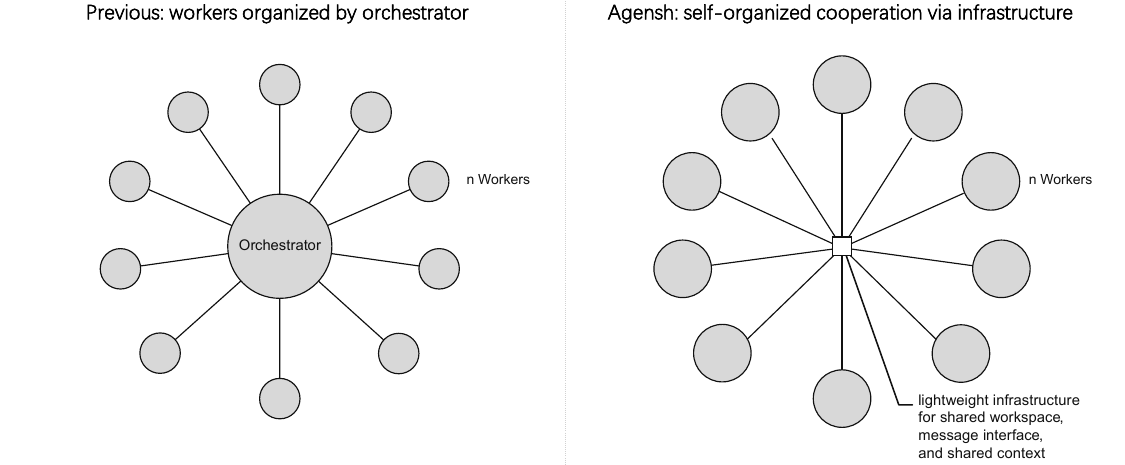}
  \setlength{\abovecaptionskip}{12pt}
  \caption{Comparison of multi-agent system. In previous work, cooperation among agent workers is
  constrained by the capacity of a central orchestrator. In Agensh, self-organized workers share
  their state and communicate their progress through the lightweight agentic organization infrastructure.}
  \label{fig:organization-comparison}
\end{figure}

\section{Introduction}
\label{sec:intro}

Single-agent systems allow large language models to perceive context, take actions through tools,
update state, and iteratively progress toward a goal \citep{yao2023react,shinn2023reflexion}.
However, their capability is bounded by one
context window, one action stream, and one memory stream. The sequential execution constraint
imposes high latency on complex real-world tasks. When we want to overcome these constraints, we
need to scale the number of agents in multi-agent systems \citep{kim2025scalingagents}.

Several pioneering harness frameworks support multi-agent systems, including Codex sub-agent,
Claude Code sub-agent and agent teams, Copilot fleet, and Kimi Agent Swarm
\citep{codex2026,anthropicSubagents,anthropicTeams,copilotFleet2026,kimi2026}. These frameworks widely adopt an orchestrator-worker
structure, in which a main agent plans, decomposes tasks, assigns them to concurrent workers, and
manages them. This structure usually requires a trained orchestrator to plan and
decompose tasks effectively \citep{kimi2026}. However, the scalability of the overall system remains fundamentally
limited by the orchestrator's capacity to manage and coordinate its workers and integrate their contributions
\citep{yang2025agentnet,anthropic2025multiagentresearch,anthropic2026multiagentsystems}.

To address this limitation, we introduce \ours{}, a scalable multi-agent organization harness
without a central orchestrator. Self-organized workers run concurrently and asynchronously, and share their
state and progress through the lightweight agentic organization infrastructure, as illustrated in
Figure~\ref{fig:organization-comparison}. The infrastructure
consists of three coordination components: a shared workspace that holds the organization's
ongoing and integrated work, a message interface that carries organization-wide announcements
and urgent direct messages, and shared context that retains peers' reusable findings and work
intentions.

\ours{} further couples an asynchronous cooperation loop based on the infrastructure for each worker.
In each loop iteration, a worker first gathers context for the shared user's goal and reads peer progress,
and claims a sub-task to carry out itself. After addressing the potential overlaps or conflicts
through messages, the worker takes action to solve the sub-task using available tools. During
this work, it updates the shared context whenever it establishes findings that may help other
workers. Finally, after its work is finished and verified, the worker merges its progress into
the shared workspace and loops back to gathering context for its next sub-task. Overall, the
multi-agent cooperation loop enables workers to turn individual progress into collected
contributions, while the infrastructure makes
accumulated work, findings, and messages accessible throughout the organization.

We test the scalability of \ours{} on ProgramBench \citep{programbench2026}, one of the most
challenging benchmarks for agentic software engineering, where an agent organization must
reproduce the behavior of reference software without Internet access given a 6h budget. We
select the five most difficult tasks, whose reference repositories comprise thousands of files,
with source code spanning millions of bytes.
In our experiments, increasing the organization from 1 to 128 agents raises the mean final
test-pass rate across the five tasks from 19.31\% to 28.78\%, an approximately 49\% relative
improvement. Larger organizations can reach comparable test-pass rates earlier, showing that
more concurrent agents can reduce the latency to a given level of performance.  On
\texttt{pandoc}, scaling from 1 to 1,024 agents raises the final test-pass rate from 33.89\% to
55.06\%. Recorded trajectories further reveal progressively
broader forms of self-organized cooperation as the organization grows, from peer coordination and multi-worker integration to
standardized workflows and specialized roles at organization scale. Overall, \ours{} reveals agent count as a new
scaling dimension for multi-agent organizations: it not only offers a practical solution for
complex tasks under hard latency constraints or time budgets, but also shows how scaling a
multi-agent organization can expand the frontier of general intelligence.

\begin{figure}[t]
  \centering
  \makebox[\linewidth][c]{%
    \includegraphics[width=1.20\linewidth]{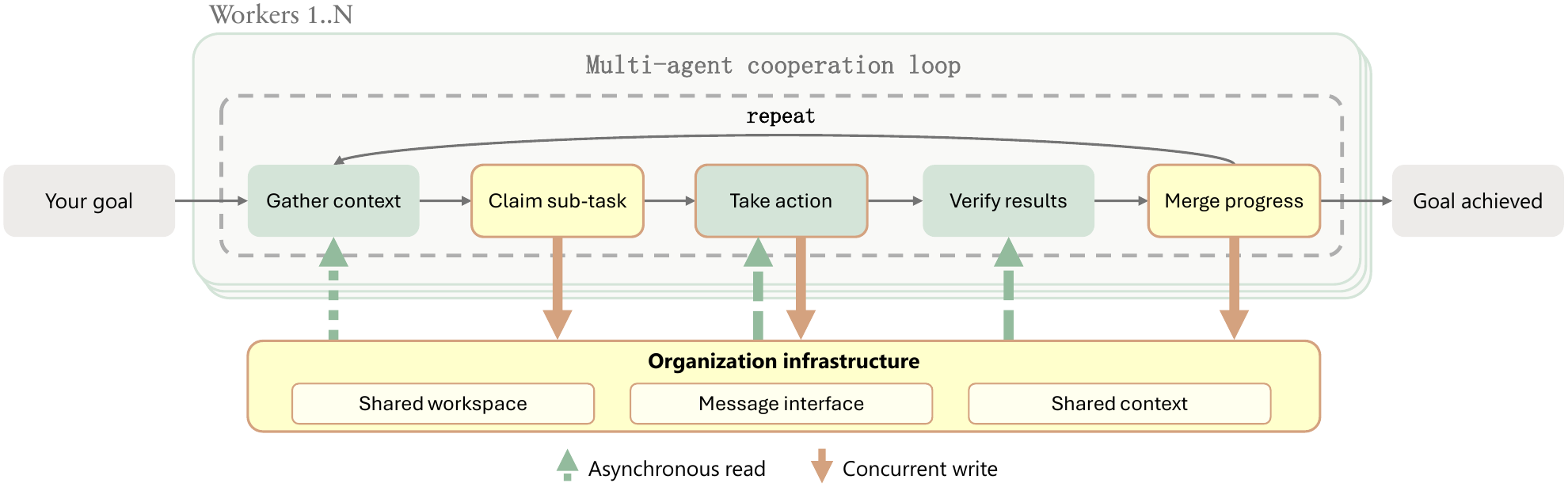}}
  \setlength{\abovecaptionskip}{6pt}
  \caption{Agensh multi-agent cooperation loop. Workers concurrently gather context, claim
  sub-tasks, take action, verify results, merge progress, in a self-organized manner, and repeat
  until the user's goal is achieved. All workers proceed asynchronously and share progress through
  the infrastructure.}
  \label{fig:loop}
\end{figure}

\section{Multi-Agent Organization Harness}
\label{sec:harness}


In this section, we introduce \ours{}, which couples a multi-agent cooperation loop with the
agentic organization infrastructure. The multi-agent cooperation loop guides all workers in turning
individual progress into verified contribution integrated with the organization's work. The
infrastructure makes accumulated work, findings and messages available across the multi-agent organization.

\subsection{Multi-Agent Cooperation Loop}
\label{sec:cooperation-loop}

The core abstraction of a harness is its loop. As illustrated in Figure~\ref{fig:loop}, each
worker in \ours{} executes the following five-step cooperation loop concurrently and asynchronously:

\begin{enumerate}[leftmargin=2em,itemsep=1pt]
  \item \textbf{Gather context.} The worker reads the shared user's goal, current state, peer progress
        and messages, and accumulated findings to understand what has been accomplished and what
        remains. It also interacts with the environment to identify and plan its next useful
        steps.
  \item \textbf{Claim sub-task.} The worker proposes a sub-task to do and announces its scope as a
        \texttt{CLAIM} appended to the \emph{shared context}. When two workers' claims overlap or conflict,
        they are encouraged to resolve the issue through \emph{direct messages}.
  \item \textbf{Take action.} The worker works locally to solve the sub-task it proposed by interacting
        with the environment through available tools. It also reports intermediate progress
        through the \emph{shared context} whenever it establishes findings that may help
        other workers.
  \item \textbf{Verify results.} The worker matches its local progress against the sub-task's acceptance
        criteria. If it does not satisfy those criteria, the worker revises its approach until the criteria are met.
  \item \textbf{Merge progress.} The worker makes its contribution available to peers by
        merging it into the \emph{shared workspace}. Then it publishes an update describing what is changed,
        the underlying idea, and verification evidence so that peers can build on the result.
        If the merge is blocked by a conflict, the worker incorporates the latest peer progress,
        resolves the conflict, and merges again.
\end{enumerate}

After merging progress in step~5, the worker goes back to step~1 and gathers context for its next sub-task again, based on newly integrated work and the latest peer updates. By proposing and claiming
their own sub-tasks, workers self-organize sub-task discovery and allocation across the organization.
They proceed asynchronously, allowing each worker to continue without waiting for
all peers to complete an iteration. As workers repeat this process, their findings and
verified contributions accumulate through the shared infrastructure, driving the organization
towards the user's goal.

\subsection{Agentic Organization Infrastructure}
\label{sec:infrastructure}
\label{sec:channels}

\begin{figure}[t]
  \centering
  \includegraphics[width=0.60\linewidth]{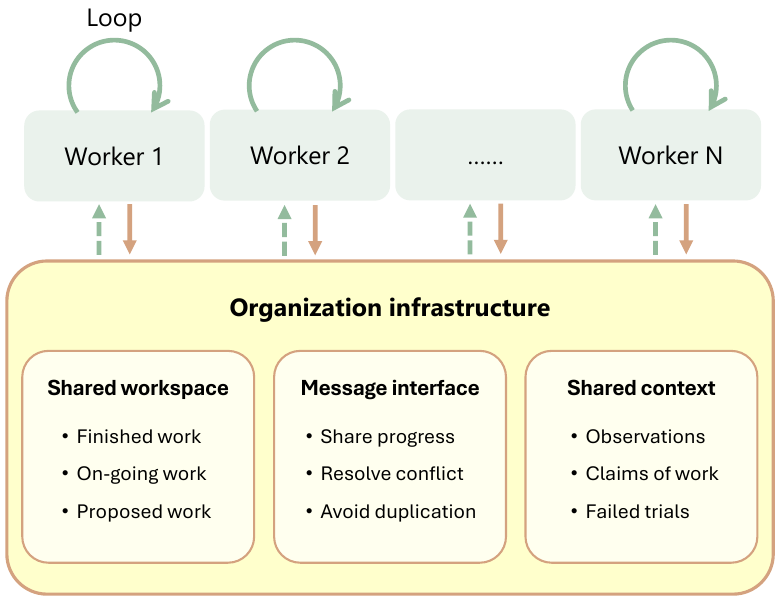}
  \setlength{\abovecaptionskip}{6pt}
  \caption{\ours{} organization infrastructure. The shared workspace holds the organization's
  ongoing and integrated work, the message interface carries organization-wide announcements and urgent
  targeted communication, and shared context retains reusable findings and claims of work.
  Together, these infrastructure components sustain a persistent, self-organized multi-agent organization.}
  \label{fig:overview}
\end{figure}

Workers in a multi-agent organization need to access one another's work, communicate about
their sub-tasks, and retain findings to guide their next steps. As illustrated in
Figure~\ref{fig:overview}, \ours{} supports three coordination
mechanisms: a \emph{shared workspace}, a \emph{message interface}, and \emph{shared context}.

\paragraph{Shared workspace.} The shared workspace is a file system intended to hold the organization's work,
including under development and integrated results. It is expected to support
concurrent writes and asynchronous reads, allowing workers to develop separate contributions
while accessing work published by peers. It should also preserve the version history and support
merging contributions, exposing the merge conflict for workers to backtrack and resolve. In \ours{}, we use a
Git platform to manage the shared workspace. Workers modify private checkouts and
branches and integrate their contributions into a main
branch. Git records the provenance of changes and detects textual merge conflicts, while
hosting issues and pull requests for all.

\paragraph{Message interface.} The message interface is intended to let workers coordinate their
ongoing work through communication, address ownership overlaps, and resolve dependency conflicts.
In the \ours{} message interface, a shared task channel carries team announcements, while direct
messages support urgent one-to-one messages. Lower-priority shared-task-channel messages are delivered at the
beginning of each loop iteration, while higher-priority direct messages are delivered at the end of each infrastructure tool call. The message interface also retains the conversation history and delivers messages asynchronously.
These functions allow workers to resolve overlapping claims, negotiate dependencies, and
request assistance in a timely manner while their peers continue to work.

\paragraph{Shared context.} The shared context is intended to retain findings
that workers can reuse and claims of work across the organization. The idea is adopted from DeLM
\citep{delm2026}. In \ours{}, workers publish concise, typed shared context entries: \texttt{OBSERVED}
behavior, confirmed \texttt{FACT}s, unsuccessful approaches recorded as \texttt{FAIL}, active
\texttt{CLAIM}s, and \texttt{PATCH\_SUMMARY} entries describing completed changes. The service
retains an append-only database with recent memory. To support longer task
horizons, we introduce a
context grep tool that lets workers search the full recorded history beyond recent memory. New entries on the shared context are
treated as higher-priority updates and will be forwarded when every other worker's next infrastructure tool call returns,
making peer findings visible across the organization.

Together, these mechanisms connect each worker's local activity to the organization's shared
state. A worker can reuse a peer's findings, address sub-task overlapping, and make an
integrated contribution within its own cooperation loop.

\subsection{Implementation}
\label{sec:implementation}

\ours{} is an organization harness layered above a single-agent harness rather than a
replacement for it. For each worker, the underlying single-agent harness owns the local agentic loop:
it maintains conversational state, invokes the model, executes tools, and produces the worker's
next response. \ours{} supplies organization-level behavior around that loop, including worker
identity and the protocols for cooperation, event routing, robust dispatch, shared-workspace
access, messaging, shared context, and recovery and liveness mechanisms. Thus, the single-agent
harness determines how one worker reasons and acts, while \ours{} determines how many workers
receive events, share state, coordinate, and repeatedly contribute to a shared goal.

The interface between the single-agent agentic loop and the multi-agent cooperation loop is intentionally kept minimal and plug-and-play. \ours{} realizes the cooperation loop through workflow instructions in each
worker's prompt rather than hard-coding the loop into the runtime infrastructure. The complete worker prompt is presented in
Appendix~\ref{app:worker-prompt}, which is exactly the same for each worker except for the worker ID. Therefore, Agens can connect to different
underlying harnesses, such as Claude Code and Copilot, through lightweight
harness-specific adapters without changing the cooperation loop, the shared services, or the
agentic loop of the underlying harness.

We implement the shared workspace with Gitea \citep{gitea2026github} and the message interface
with Mattermost \citep{mattermost2026github}. The shared context follows the core idea of DeLM
\citep{delm2026}, but we adapt its tool formats and worker instructions.
Appendix~\ref{app:implementation} further describes how repository events, messages, and shared
context are delivered to workers.

\begin{figure}[t]
  \centering
  \includegraphics[width=\linewidth]{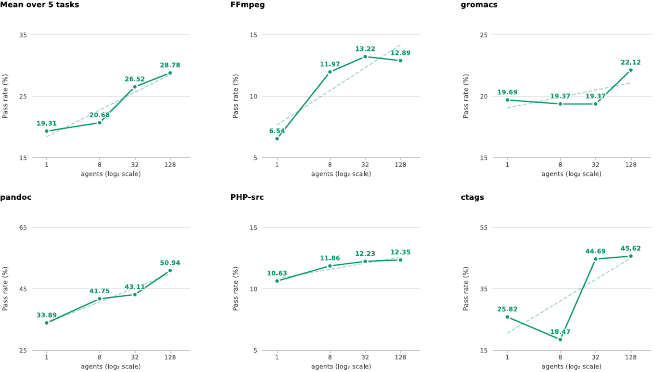}
  \caption{Scaling from 1 to 128 agents on the five hardest ProgramBench tasks
  \citep{pandoc2026github,ffmpeg2026github,gromacs2026github,php2026github,ctags2026github}
  under 6h budget.
  Mean final test-pass rate rises from 19.31\% for 1 agent to 28.78\% for 128
  agents.}
  \label{fig:scaling}
\end{figure}

\section{Scaling Multi-Agent Organization to 1,024 Agents}
\label{sec:scaling}

\begin{table}[b]
  \centering
  \caption{Reference repository size of the five hardest tasks in ProgramBench.}
  \label{tab:repository-scale}
  \small
  \begin{tabular}{llrrr}
    \toprule
    Task & Commit & Files & Lines of code & Code size (MB) \\
    \midrule
    \texttt{FFmpeg} \citep{ffmpeg2026github} & \href{https://github.com/FFmpeg/FFmpeg/commit/360a4025fb2582d52d871ea2129d6b659598bb49}{\texttt{360a402}} & 10,090 & 1,549,005 & 75.51 \\
    \texttt{gromacs} \citep{gromacs2026github} & \href{https://github.com/gromacs/gromacs/commit/665ea4ca703128d057f978d9d2f99871aa816618}{\texttt{665ea4c}} & 8,982 & 815,539 & 47.30 \\
    \texttt{pandoc} \citep{pandoc2026github} & \href{https://github.com/jgm/pandoc/commit/5caad90fc44669ced0abdd7a890108a995689a90}{\texttt{5caad90}} & 2,767 & 104,336 & 4.86 \\
    \texttt{PHP-src} \citep{php2026github} & \href{https://github.com/php/php-src/commit/c8912639008a89a195a258fc8cb924b5763c6766}{\texttt{c891263}} & 26,266 & 2,812,757 & 119.76 \\
    \texttt{ctags} \citep{ctags2026github} & \href{https://github.com/universal-ctags/ctags/commit/243595eebc1178ac04695524ee5f8356cc28d480}{\texttt{243595e}} & 7,313 & 246,228 & 9.15 \\
    \bottomrule
  \end{tabular}
\end{table}

We use ProgramBench \citep{programbench2026} to test whether scaling a multi-agent organization
can extend to a new frontier of agentic software engineering on complex long-horizon tasks.
ProgramBench requires agents to reconstruct reference software's behavior from scratch given
a 6h budget, with Internet access disabled to prevent retrieval of existing implementations. We
test Agensh on five particularly demanding systems: \texttt{FFmpeg} \citep{ffmpeg2026github},
\texttt{gromacs} \citep{gromacs2026github}, \texttt{pandoc} \citep{pandoc2026github},
\texttt{PHP-src} \citep{php2026github}, and \texttt{ctags} \citep{ctags2026github}. These are the five most difficult tasks among
ProgramBench's 200 instances, as measured by the mean test-pass rate of state-of-the-art models. The systems span multimedia processing, molecular
simulation, document conversion, language interpretation, and code indexing. Their reference
repositories contain thousands of files and hundreds of thousands to millions of lines of code,
as collected in Table~\ref{tab:repository-scale}, making them a demanding test of long-horizon
multi-agent cooperation.

\subsection{Main Results}

Figure~\ref{fig:scaling} demonstrates that agent count is a scaling dimension for multi-agent
organizations. With the same agent model, GPT-5.6-sol (high), underlying single-agent harness, Copilot, and 6h budget, the mean
final test-pass rate across the five tasks increases from 19.31\% with 1 agent to 20.68\%,
26.52\%, and 28.78\% with 8, 32, and 128 agents, respectively. Scaling from 1 to 128 agents
yields a gain of 9.47 percentage points, or an approximately 49\% relative improvement in the
average score. Across the five tasks, final scores show a generally increasing trend as the
agent organization grows, indicating that continuously increasing the number of cooperating workers can consistently and substantially improve the quality of complex software reproduction over long time horizons.

\begin{figure}[t]
  \centering
  \includegraphics[width=\linewidth]{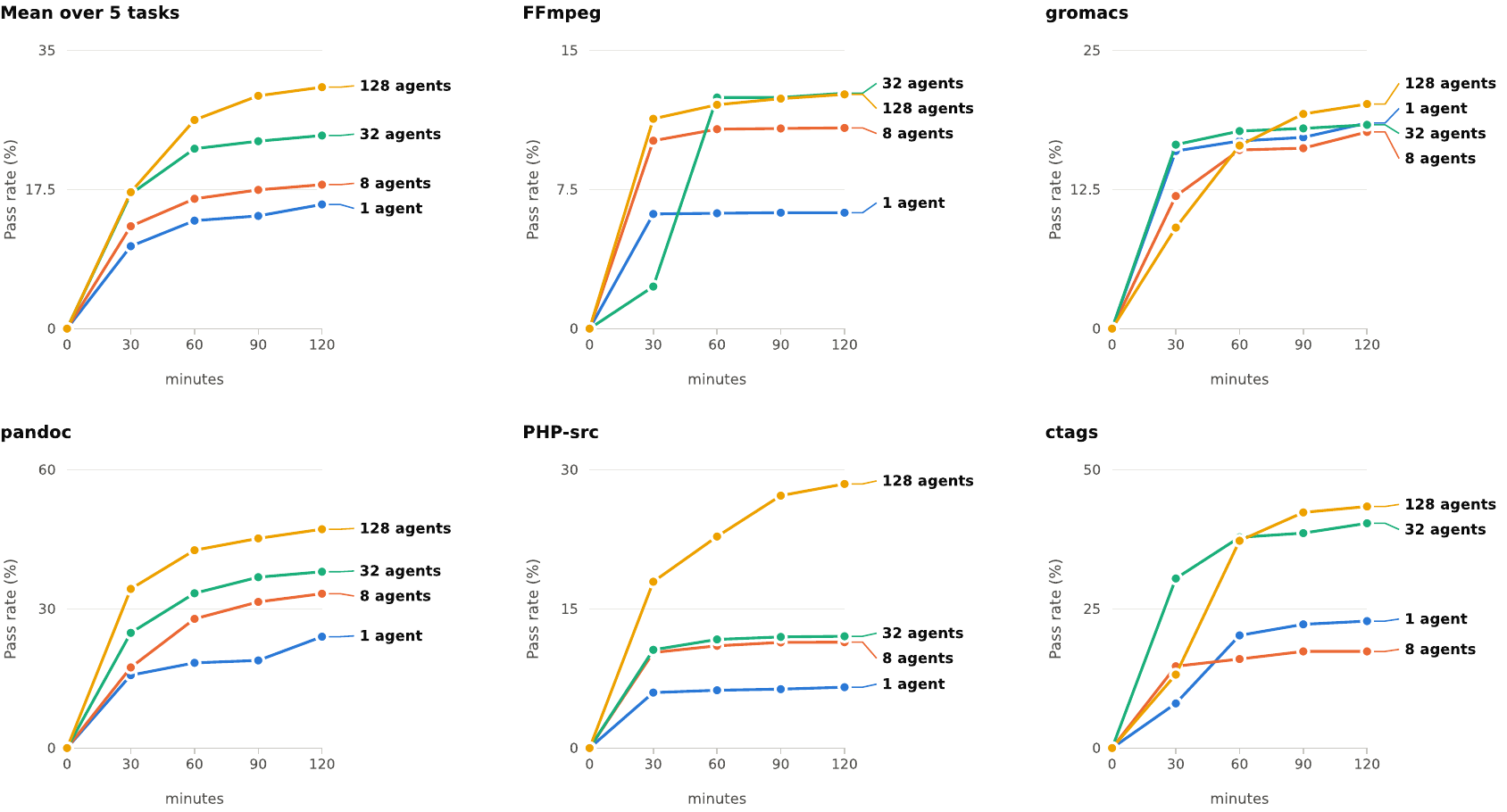}
  \caption{Test-pass rates over time with 1 to 128 agents on the five hardest ProgramBench tasks
  during the first 2h of the 6h runs. Larger organizations generally achieve higher test-pass
  rates earlier.}
  \label{fig:trajectories}
\end{figure}

Figure~\ref{fig:trajectories} further demonstrates that more concurrent agents can also reduce the latency to
achieve the same score. During the first two hours, larger organizations reach comparable
test-pass rates earlier. For example, on \texttt{pandoc}, 128 agents exceed a 30\% test-pass
rate at the 30-minute checkpoint, while 32 and 8 agents first exceed that threshold at the
60- and 90-minute checkpoints, respectively. The single-agent run remains below this threshold
throughout the first two hours. Increasing agent count can therefore shorten the time needed
to reach a given level of task performance, in addition to improving the final score.

Figure~\ref{fig:headline-scaling} extends this scaling to 1,024 agents on
\texttt{pandoc}. Under the same 6h budget, the final test-pass rate rises from 33.89\%
with 1 agent to 50.94\% with 128 agents and 55.06\% with 1,024 agents. The largest organization
improves on the 128-agent result by 4.12 percentage points and the single-agent result by
21.17 percentage points. These results show that Agensh continuously scales self-organized cooperation among
more than a thousand agents.

With the underlying model and harness held fixed, these experiments show that increasing the
number of workers can improve both the quality and speed of software reproduction. The gains
across five demanding tasks, together with the extension to 1,024 agents on \texttt{pandoc},
support agent count as a new scaling dimension for multi-agent organizations. They provide evidence
that organizational intelligence can grow through scaling self-organized cooperation on complex, long-horizon
work.

\subsection{Self-Organized Cooperation Emerges at Scale}
\label{sec:self-organized-cooperation}

Beyond the performance gains, the recorded trajectories reveal how workers organize their own
cooperation. All workers follow the same loop and receive the same prompt except for worker IDs,
but new forms of self-organized cooperation progressively emerge as the organization grows.

\paragraph{8 agents: coordinating implementation with peers.}
Workers can agree on a concrete technical interface and independently implement components that
conform to it. In \texttt{gromacs}, workers announced a module interface and independently
implemented command modules that followed it.
Workers can also discover and resolve overlapping claims themselves. In
\texttt{FFmpeg}, one worker changed its scope and took on complementary work
after discussing the overlap issue with a peer.

\paragraph{32 agents: managing integration across multiple workers.}
Multiple workers can jointly work on a technical contribution. In \texttt{PHP-src},
several peers initially approved a contribution. Another worker found a concrete counterexample,
so the earlier approval was withdrawn, and the author fixed the problem. Peers then reviewed the
contribution again, and a peer merged it. Workers also became more active in
managing integration, and a broader set of peers participated in the communication.

\paragraph{128 agents: self-organized specialization and workflow standardization.}
Specialization emerges at this scale. Workers are choosing reviewers based on relevant prior
experience and reuse these review relationships over time. They can also transfer responsibility
for integrating a contribution to a peer who resolves conflicts, validates the combined work,
and merges it.

Workers can negotiate, follow, and reuse standardized self-organized workflows. In \texttt{pandoc}, two
workers established a standardized integration protocol: the author updated and tested a branch,
then sent its commit hash for peer validation and merge. This protocol was later reused by other
workers. After some failures, workers further revised the cooperation protocol: they agreed to
give the peer permission to perform the entire update, test, check, and merge cycle. The peers
explicitly accepted and carried out this revised procedure for new pull requests.

\paragraph{1,024 agents: role specialization at organization scale.}
At this scale, multiple workers take on the same specialized roles or develop expertise in the
same technical area. In \texttt{pandoc}, multiple workers served as integrators. A worker could
contact several candidate integrators, select the first valid responder, cancel the other
requests, and hand over the code to be integrated only after this selection. Workers specializing in
the same technical area can also recover and take over the work after another worker's attempt
fails. This strengthens the organization’s robustness by avoiding dependence on any individual worker.

These observations complement the scaling results by showing how new forms of cooperation emerge
while the number of agents scales. As the organization grows, cooperation becomes broader in scope, extending from
coordinating implementation to managing integration, standardizing workflows, and developing
specialized roles at organization scale. Surprisingly, these forms of cooperation are entirely
self-organized: workers themselves choose collaborators, divide responsibilities, and establish
and revise their workflows through peer interactions. This further illustrates the scaling
potential of self-organized multi-agent systems.

\section{Related Work}
\label{sec:related}

\paragraph{Multi-agent organization and cooperation harnesses.} Existing harnesses organize multi-agent systems through varying approaches to task assignment, communication, and integration. Claude Code supports persistent sub-agents and asynchronous agent
teams, but the collaboration is still organized around a fixed lead session that spawns
other agents, manages shared tasks, and coordinates with inter-agent messages
\citep{anthropicSubagents,anthropicTeams}. Codex supports sub-agent workflows with delegated parallel workers whose results are routed back to the parent for consolidation \citep{codex2026}. Kimi Agent Swarm trains an orchestrator to
decompose tasks and schedule sub-agents concurrently for lower latency, but the harness
remains an orchestrator-worker parallelization scheme \citep{kimi2026}. We study a multi-agent
organization structure in which self-organized workers share responsibility for discovering tasks, coordinating dependencies, and integrating results over persistent work cycles.

\paragraph{Shared context and decentralized coordination.} DeLM combines asynchronous workers, a task queue, and shared context so that agents can build on
accumulated progress without routing every update through a central controller \citep{delm2026}.
Existing collaboration tools also support coordination among agents. ChatCollab places human and
AI participants in Slack as peers that coordinate roles, requests, and progress
\citep{chatcollab2024}. SlackAgents embeds AI agents into Slack workspaces for communication and
task orchestration \citep{slackagents2025}. For the shared workspace, Carlini's compiler-building
experiment coordinates 16 Claude agents through a Git repository and task-lock files, with
workers selecting tasks and merging changes without an orchestration agent \citep{carlini2026}.
STORM reframes multi-agent coding as a state-management problem, replacing isolated multi-agent
worktrees with a shared workspace that enforces local state consistency and detects stale or
conflicting edits at write time \citep{storm2026}.

\section{Conclusion}
\label{sec:conclusion}

We introduced \ours{}, a scalable multi-agent organization harness that couples a multi-agent
cooperation loop with lightweight agentic organization infrastructure. Through a shared workspace, a
message interface, and shared context, self-organized workers discover and claim sub-tasks,
exchange findings, and integrate contributions while proceeding concurrently and asynchronously.
This design enables individual progress to accumulate into shared work and knowledge without
relying on a central orchestrator to allocate tasks and manage workers.
In our experiments, with the same agent model GPT-5.6-sol (high) and 6h budget, scaling from
1 to 128 agents on the five hardest
ProgramBench tasks raises the mean final test-pass rate from 19.31\% to 28.78\%, an approximately
49\% relative improvement. Larger organizations can reach comparable test-pass rates earlier,
reducing the latency to a given level of performance. In \texttt{pandoc}, scaling to 1,024 agents
raises the final test-pass rate from 33.89\% with 1 agent to 55.06\%. Worker trajectories further
reveal the forms of self-organized cooperation broadening with scale, from peer coordination and multi-worker integration
to standardized workflows and organization-scale role specialization. These results demonstrate
the number of agents as a new scaling dimension for multi-agent organizations. Through \ours{}, scaling
cooperation not only offers a practical approach to complex tasks under hard latency constraints
or time budgets, but also opens a path to expand the frontier of general intelligence.

\clearpage
\bibliographystyle{unsrtnat}
\bibliography{references}

\clearpage
\appendix
\section{Worker Prompt}
\label{app:worker-prompt}

The 1,024-agent configuration supplies the following prompt to
\texttt{pbench-team1-worker1}. The prompt for every other worker is exactly the
same except for the worker ID.

\begin{Verbatim}[
  fontsize=\scriptsize,
  breaklines=true,
  breakanywhere=true,
  breaksymbolleft={}
]
# Hello

You are pbench-team1-worker1.

## Where you are

You work with 1024 peers. Work is tracked in **gitea**; you talk on
**mattermost**; you share findings on the **shared context board**.

- **gitea** -- issue #1 is the task. You open your own issues for the areas you take, and PRs
  carry the work. Use the in-container address **`http://gitea:3000`** for everything: clone,
  fetch, push, curl, API. The `clone_url` / `html_url` gitea returns are external addresses
  (`http://gitea.localhost:<port>`) and are **unreachable from your container** -- substitute
  `http://gitea:3000` whenever you see one.
- **mattermost** -- channel `pbench-task`, where the team talks. Be brief and concrete.
  Handles are `pbench-team1-worker1..1024`; `@worker1` matches no account, so
  use the full one. A **direct message** on the same server reaches a peer **inside their
  current turn**, so it is the fast way to settle a collision -- and it interrupts what they are
  doing. Send a direct message when two of you are about to edit the same thing; use the
  channel for everything else.
- **shared context board** -- an append-only log of short, verified notes from every agent. It is
  rendered into your prompt each turn under `==== SHARED CONTEXT ====`, and anything a peer
  writes after that reaches you mid-turn, appended to whatever tool result you were waiting on.
  Publish with `board_write` the moment you establish something -- `OBSERVED` for noticeable
  reference behaviour, `FACT` for what you confirmed, `FAIL` for a hypothesis you falsified (the
  highest-value entry: it stops peers spending budget on it), `PATCH_SUMMARY` as
  `files= | idea= | evidence=` when you fix something, and `CLAIM` while you are working so
  peers pick a different angle. Entries are capped (100 chars, 300 for `PATCH_SUMMARY`); put the
  long version in `detail` and peers can `board_unfold` it. Do not re-derive a peer's `FACT`,
  and do not retry something recorded as `FAIL`. `board_read` returns the full recent board
  (truncated to the 2000 latest entries) while `board_grep` returns entries containing your
  keywords across the complete history. You can grep multiple keywords: ',' is OR and '&' is
  AND (case-insensitive), so `a&b,c&d` means (a AND b) OR (c AND d).

## The team

1024 coders, all equal. You claim what to build, you build it, you merge it, and
you choose the next thing. Your peer coders are working simultaneously: sync your codebase with
the latest `main` branch before merging your own code to `main`, and sync up with peers to avoid
duplicated work. Always focus on reproducing the reference; reach for the coordination tools
only when your information can greatly impact others' work.

# Current task: ProgramBench reverse rebuild

## Goal

You are given a **pre-compiled reference program** (binary), mounted at `/opt/pbench/executable` in every container.
Task: **rebuild from scratch a complete codebase that reproduces its behavior**, so that it can be compiled and pass the hidden tests.

- You **can run** this binary to observe its behavior: `/opt/pbench/executable --help`, feed various inputs and inspect the output,
  compare boundary cases. This is the **only** information source for understanding the target behavior (plus docs / assets in the repo).
- You **cannot read its bytes**: it is `root:root 0111`, `cat` / `od` / `objdump` will get Permission denied.
  This is a rule, not a fault -- **do not try to decompile or dump it**; that is cheating, and even if you try, you can't read it.

## Iron rules (violating any = the entire task is voided)

1. **Fully autonomous**: the whole team defines requirements itself, breaks down work itself, decides itself, and finishes itself. There is no external input;
   do not wait for instructions, do not ask on mattermost "what to do next" -- the answer is always: **push the rebuild forward**.
2. **No internet**: the whole cluster is offline. `git clone` of the original project source, `pip install`, `npm i` of remote packages
   **will all fail**. Do not waste time trying to fetch the original source over the network -- that is both impossible and cheating. Only the toolchain
   already available in the container (gcc / make / python / node etc. installed locally) is usable.
3. **Hard delivery constraint**: the repo **root directory** must have an executable `compile.sh`; running it must produce an executable file
   named `./executable` at the **root directory**. The evaluator will `tar` your repo HEAD -> unpack -> **delete any committed `./executable`** -> run `./compile.sh`
   -> run the hidden tests using the produced `./executable`. `compile.sh` must therefore create `./executable`, not just `chmod` it.
   **No root-directory compile.sh = 0 points**.

## The first step is always

**Run `/opt/pbench/executable` and observe what it does.** Every rebuild starts from understanding its actual behavior.

## Working with the tools

**Events.** You are event-driven. When something relevant happens, the environment sends you
one message:

    [event]
    event_id=...
    source=...     <- gitea / mattermost / ...
    kind=...
    observed_at=...
    summary: ...   <- what happened and where to find it

It reports facts; deciding what to do about them is yours. **A direct message and a peer's new
board entry reach you mid-turn**, appended to whatever tool result you were already waiting on.
Everything else queues and arrives when your turn ends.

A board entry arriving mid-turn provides information:

- `FAIL` -- stop if you are doing that thing.
- `FACT` / `OBSERVED` -- use it.
- `CLAIM` on what you are building -- a collision; settle it by direct message.
- any other `CLAIM` -- keep building what you are on.

**The work cycle.** Each step is finished by the person doing it.

1. Run `/opt/pbench/executable` to observe the unexplored areas to build.
2. Announce the area you are taking as a `CLAIM` on the shared context board.
3. Run `/opt/pbench/executable` on that area under `ulimit -v 67108864` and write code that
   reproduces what you observe, based on current `main`. You do not need to sync code with
   others during implementation.
4. Check your work by running the reference and your build on the same inputs and comparing.
5. Push, open a PR, and merge to `main`. If the merge is refused, merge the latest `main` into
   your branch, re-check your work against the reference, and push again.
6. Say what you finished as a `PATCH_SUMMARY` (`files= | idea= | evidence=`) on the board,
   then go back to 1.

Commit everything the build needs: the grader builds from a fresh clone of `main` and does not
invoke your binary from the repo root, so resolve data paths with `dirname(realpath(argv[0]))`.

**Anything else.** MCP, git CLI, shell, curl, writing files -- whatever is at hand is fine.
If the tool you have is not enough, look for one; if there is none, build one.
\end{Verbatim}

\section{Implementation Details}
\label{app:implementation}

\subsection{Event-Driven Runtime}
\label{sec:runtime}

Each worker is driven by events rather than by a centrally scheduled sequence of steps. Its
paired router listens for repository activity from Gitea and messages from Mattermost. For Gitea, a server-sent event triggers a fetch of notifications newer than the
last processed notification; for Mattermost, a live connection delivers messages and a reconnect
catch-up retrieves messages missed while disconnected. When
new activity arrives, the listener writes the event to the worker's durable queue and wakes the
dispatcher. Each event has a unique identifier for
deduplication.

Each dispatcher wake starts from the worker's queued events. The dispatcher orders and combines
pending events into the next prompt, delivers it to the worker's persistent session, and
marks the events as sent when the turn completes. Events that arrive during a turn remain queued
for a subsequent turn. Failed deliveries are retried after progressively longer waiting periods;
after a router restart, unfinished events return to the queue and the worker's prior session is
restored when possible. If a worker remains idle for 10 minutes, an idle detector sends a prompt
asking it to continue working. If a completed turn produces no shared context entry, a follow-up
prompt reminds the worker to publish useful findings.

\subsection{Shared Context Delivery}
\label{app:delivery}

Shared context reaches workers through two delivery paths. At the beginning of a new turn, the router appends the queued
workspace notification, arriving messages, and the recently shared context to the prompt. During an active turn, a wrapper around any infrastructure tool call
appends pending direct messages and new shared context entries to the tool results returned.
This path operates on MCP tool returns; it
does not interrupt an arbitrary running process.

\section{Experiment Details}
\label{app:evaluation}

\paragraph{Task selection.} We select the five hardest tasks based on the official ProgramBench extended results at \url{https://programbench.com/extended/}, where task difficulty is determined by the average pass rate of state-of-the-art models.

\paragraph{Agent configuration.} All reported configurations use \texttt{gpt-5.6-sol} with high reasoning effort and the same harness, worker protocol, and evaluation configuration. We use Copilot as the underlying single-agent harness, with a maximal input token limit of 272,000 and a maximal output token limit of 128,000.

\paragraph{Experiment setup.} We follow ProgramBench to prepare the agent containers, set the total time budget to 6 hours, run the official evaluation, and report the canonical-kept pass rate.

\paragraph{Large-scale deployment.} To reduce scope contention, we stagger agent activation in all experiments, launching one agent every 30 seconds during the first hour and one every 3 seconds thereafter. Let \(t\) denote the time at which the first agent is activated. At \(T=t+6\,\mathrm{h}\), all agents are terminated and the submission is exported.

\noindent At \(T-45\,\mathrm{min}\), we send the following reminder:

\begin{Verbatim}[
  fontsize=\scriptsize,
  breaklines=true,
  breakanywhere=true,
  breaksymbolleft={}
]
Stop dispatching new features. Ensure the root build works and existing work is merged. Land open PRs; if a PR cannot be merged, say so and move on.
\end{Verbatim}

\noindent At \(T-5\,\mathrm{min}\), we send a second reminder:

\begin{Verbatim}[
  fontsize=\scriptsize,
  breaklines=true,
  breakanywhere=true,
  breaksymbolleft={}
]
Merge anything ready, then stop. Confirm bash compile.sh && ./executable works on the default branch and post the final state.
\end{Verbatim}

In the 1,024-worker experiments, we distribute workers across 16 nodes, with 64 agents per node.

\paragraph{Single-agent baseline.} Since in practice, a single-agent run can rarely sustain the full 6-hour runtime, we add a stop hook to the single-agent baseline with the following prompt:

\begin{Verbatim}[
  fontsize=\scriptsize,
  breaklines=true,
  breakanywhere=true,
  breaksymbolleft={}
]
There is always new work -- do not be bounded by the previous plan's scope: pick up /opt/pbench/executable, run it against untried inputs, and derive fresh tasks from what interaction with it reveals. Let's make the 6h fulfilled.
\end{Verbatim}

This is paired with an idle detector for the multi-agent organization that sends the following
prompt after an agent has been idle for 10 minutes:

\begin{Verbatim}[
  fontsize=\scriptsize,
  breaklines=true,
  breakanywhere=true,
  breaksymbolleft={}
]
There is always new work -- do not be bounded by the previous CLAIMs' scope: pick up /opt/pbench/executable, run it against untried inputs, and close the gap between reference and the team's implementation.
\end{Verbatim}

\end{document}